\documentclass{article}

\usepackage[preprint]{bdu_2024}

\usepackage[utf8]{inputenc} 
\usepackage[T1]{fontenc}    
\usepackage{hyperref}       
\usepackage{url}            
\usepackage{booktabs}       
\usepackage{amsfonts}       
\usepackage{nicefrac}       
\usepackage{microtype}      
\usepackage{xcolor}         

\usepackage{algorithm}      
\usepackage{algpseudocode}  
\usepackage{amsmath}        
\usepackage{mathtools}

\usepackage{cleveref}

\usepackage{natbib}         

\usepackage{wrapfig}

\usepackage{array}
\usepackage{caption}

\newtheorem{condition}{Condition}

\title{Predicting Electricity Consumption with Random Walks on Gaussian Processes}

\author{%
  Chloé Hashimoto-Cullen \\
  University College London\\
  \href{mailto:chloe.hashimoto-cullen.23@ucl.ac.uk}{chloe.hashimoto-cullen.23@ucl.ac.uk}
  \And
  Benjamin Guedj \\
  University College London and Inria \\
  \href{mailto:b.guedj@ucl.ac.uk}{b.guedj@ucl.ac.uk}
}

\begin{document}

\maketitle

\begin{abstract}
We consider time-series forecasting problems where data is scarce, difficult to gather, or induces a prohibitive computational cost. As a first attempt, we focus on short-term electricity consumption in France, which is of strategic importance for energy suppliers and public stakeholders. The complexity of this problem and the many levels of geospatial granularity motivate the use of an ensemble of Gaussian Processes (GPs). Whilst GPs are remarkable predictors, they are computationally expensive to train, which calls for a frugal few-shot learning approach. By taking into account performance on GPs trained on a dataset and designing a random walk on these, we mitigate the training cost of our entire Bayesian decision-making procedure. We introduce our algorithm called \textsc{Domino} (ranDOM walk on gaussIaN prOcesses) and present numerical experiments to support its merits.
\end{abstract}

\section{Introduction}
\label{Introduction}

Forecasting time series is at the centre of machine learning (ML). We focus on problem settings where we might have sparse data, limited compute capacity or unseen scenarios. We instantiate this problem in the setting of short-term electricity consumption prediction, and focus on doing this at the scale of France. For energy suppliers and public stakeholders, the necessity is to be able to predict consumption even in extreme events, such as a heat or cold wave, which can lead to large variations in consumption, possibly at a fairly high granularity.

These scenarios are also characterised by the fact that they do not have as much data as more classical time series forecasting scenarios such as stock price modelling, and therefore deep learning methods which have been a huge part of the recent artificial intelligence (AI) boom might not be as appropriate as they are unable to forecast as efficiently when there is little training data. 

For time series forecasting, GPs are well-adapted as they natively quantify uncertainty. However, their performance is indexed on the size of the training data and they are computationally expensive to train. Recently, Few-Shot Learning (FSL) for time series prediction has gained attention both from theoretical \citep{iwata2020few} and applied perspectives \citep{xu2024automated}, to help mitigate the costs of training. This work is the start of a series of analyses on French regional short-term electricity consumption. We take a FSL approach to training GPs, using a set of GPs trained on synthetic data in a first instance, with the natural next step being with actual electricity consumption data, available at \url{https://www.rte-france.com/en/eco2mix/download-indicators}.

We describe our algorithm, called \textsc{Domino}, in \Cref{Methodology} and in \Cref{Experiments}, we illustrate its performance on a synthetic dataset. We briefly discuss preliminary results and lay down ideas for future works, with the aim to use this present work as a stepping stone towards broader time series forecasting problems where data is scarce, difficult to gather or induces a prohibitive cost.

\section{Methodology}
\label{Methodology}

Due to the stochastic nature of the underlying phenomenon and its periodicity, we use GPs which quantify uncertainty and handle unseen scenarios. We refer to \citet{rasmussen_gaussian_2006} for a complete reference on GPs. 


\paragraph{Notation.}
We adopt the following conventional notations to define our problem statement. We model $I$ \emph{time series}, where $i \in \{ 1, 2, \ldots, I\}$ is an indicator of the time series used as a subscript. The time series all have $N \in \mathbb{N}$ entries. The inputs are vectors $\mathbf{t_i} = [ t_{i1}, t_{i2}, \ldots, t_{iN} ] \in \mathbb{R}^N, \forall i \in \{ i \} _{i=1}^I$. The outputs are $\mathbf{y_i} = [ y_{i1}, y_{i2}, \ldots, y_{iN} ] = y_i(\mathbf{t}) = [ y_i(t_1),  y_i(t_2), \ldots, y_i(t_N)] \in \mathbb{R}^N$.

For each time series $i \in \{ i \}_{i=1}^{I}$, we look for the function $f_i : \mathbf{t} \mapsto \mathbf{y_i} + \epsilon_i(\mathbf{t})$, where $\epsilon_i \sim \mathcal{N}(0, \sigma ^2)$ with $\sigma \in \mathbb{R}^N$. To share knowledge between time series, we leverage an algorithm whose tasks share a common mean.

\subsection{Existing work: the MAGMA algorithm}
\label{MAGMA}

The Multi-tAsk GPs with common MeAn (MAGMA) algorithm \citep[introduced by][]{leroy_magma_2022} shares knowledge between time series which are modelled by the same GP - this predicts unseen time series by using the common mean, which saves training resources and enables a more accurate prediction. The model is trained with an EM algorithm. Starting from an initialisation, in turn the hyperparameters given a distribution (E-step) and the distribution given the hyperparameters (M-step) are optimised. The optimisation can be made from the values learned at the previous step.


We refer to \citet{leroy_magma_2022} for full explanations of the algorithm and its proof. The MAGMA algorithm is implemented in the \texttt{MAGMAClustR} package (\url{https://arthurleroy.github.io/MagmaClustR/}). Whilst MAGMA is a powerful predictor, it is computationally expensive. In sparse computational resource settings, this limits its applications. We look to sample the GPs output by the algorithm to transfer knowledge to unseen time series with a more frugal approach, \emph{i.e.}, by leveraging much less data.

\subsection{Our Algorithm: a Random Walk on Gaussian Processes (\textsc{Domino})}
\label{RoW GP}

The \textsc{Domino} (standing for ranDOM walk on GaussIaN PrOcess) algorithm takes the output of the MAGMA model and samples these GPs. A random walk switches between the sampled time series at each time point, following a probability for each time series. After each walk, the random walk's performance with respect to each sampled time series is evaluated. Given this, and how often each time series has been sampled during the random walk, the performance and weights of the time series are updated. Until a maximum number of epochs is reached, until the random walk and sampled time series have a Kullback-Leibler (KL) divergence which is lower than a chosen $\delta$, or until a maximum number of samples over the $\delta$ threshold have a difference to the threshold with a lower standard deviation than the standard deviation of the in lying time series points and $\delta$, the walk is repeated with the updated performances serving as a new probability at each epoch.

\paragraph{Notation.} We superscript $w$ the current epoch to identify the information which is specific to it. The performance of the sampled time series at the current epoch is $p^w$. The random walk for the current epoch is $\mathbf{y}^w \in \mathbb{R}^N$. The time series sampled at each stage of the random walk for the current epoch is $\mathbf{z}^w \in \mathbb{R}^N$. Let $P(f_i)$ be the performance of the function $f_i$.

We establish the following condition to end the random walk.
\begin{condition}\label{cond}
Let $\delta$ denote a tolerance parameter (the largest divergence between any sample and the \textsc{Domino} at any point), and let $W$ be the  largest number of epochs possible such that at least one of the three following statement holds true.
\begin{enumerate}
    \item $\forall i \in I$, $\mathrm{KL}(\mathrm{\textsc{Domino}}||\mathcal{GP}_i) \leq \delta$: all time series' KL divergence from the \textsc{Domino} is under $\delta$;
    \item For $a \in \mathbb{N}$ where $a \ll N$, there are $a$ points of all the time series whose values difference to the $\delta$ threshold have a standard deviation lower than that of the in-lying points and their difference to the $\delta$ threshold;
    \item $w \geq W \in \mathbb{N}$: the maximum number of epochs set is reached.
\end{enumerate}
\end{condition}

\paragraph{Initialisation.} Given a GP, sample the $I$ samples to walk on, set the maximum number of epochs $W$, choose the KL divergence threshold $\delta$, the maximal number of outliers $a$ and $\lambda \in \mathbb{R}_0$ the regularisation constant. Without prior knowledge, the starting performance of each sample is $p^0 = \{ p_i^0 \}_{i = 1} ^ I = \frac{1}{I}$.

\begin{algorithm}
	\caption{\textsc{Domino}.} 
	\begin{algorithmic}[1]
        \While {\Cref{cond} is False}:
            \State Initialise $g$ with a random draw from $\mathcal{I} = \{ 1, 2, \ldots, I \}$ whose hyperparameters follow a categorical distribution with hyperparameters given by:

            \begin{math}
                \bigl\{ \frac{e^{\lambda * i}}{\sum_{k = 1}^{I}e^{\lambda * k}} \bigr\}_{i = 1}^I
            \end{math}
            \State Set $ n = g $ and use the sample from the time series drawn above.
            \State Set $\mathbf{y}^w(t_1) = \mathbf{f_g}(t_1)$ the first time step using the drawn sample and $\mathbf{z^w}(t_1) = g = n$ the randomly drawn sample time series for the first step of the random walk.
            \For {$t_n \in {t_2, \ldots, t_N}$, with a probability of $p(f_i(t_n)) = \frac{e^{\lambda * i}}{\sum_{k = 1}^{I}e^{\lambda * k}} \forall i \in \mathcal{I}$:}
                \State Set $\mathbf{y}^w(t_n) = \mathbf{f_i}(t_n)$ the next step in the random walk;
                \State Set $\mathbf{z}^w(t_n) = i \mbox{ from } \mathbf{y_i}$ the time series for the step.
            \EndFor
            \For {each time step $\mathbf{y}^w = y^w ( \mathbf{t_n} ) = \{ y^w ( \mathbf{t_1} ), y^w ( \mathbf{t_2} ), \ldots, y^w (\mathbf{t_N})  \} $ of the random walk, evaluate it against the $I$ time series:}
                \State Let $\mathbf{M}^w = \{ P(f_1, y_1), \ldots, \{ P(f_I, y_I) \} \} = \{ P(f_i, y_i) \}_{i = 1}^I$ be the performances of the time series.
                \State Let $m^w = \frac{1}{I} = \sum_{i=1}^I M_i^w$ be the average of all performances across time series.
                \State Update $\mathbf{p}^w = \{ p_i^w \}_{i=1}^w$ with $\mathbf{z_i}$: set 
                
                $$\mathbf{p_i}^w = \frac{\prod_{a=1}^{w-1} \exp \bigl( \frac{1}{2} - \frac{|i \in z_a(\mathbf{t_n})|}{I}\bigr) * \mathbf{M}_i^w}{\sum_{i=1}^{I} \bigl( \prod_{a=1}^{w-1} \exp \bigl( \frac{1}{2} - \frac{|i \in z_a(\mathbf{t_n})|}{I}\bigr) * \mathbf{M}_i^w\bigr) }. $$
                
                \State Store the $\mathbf{p_w}$ performance values, $\mathbf{m^w}$ average performance across all time series, $\mathbf{M^w}$ time series performances, $\mathbf{z^w}$ steps from the random walk and $\mathbf{y}^w$ values from the random walk. for the $w^{th}$ epoch.
             \EndFor
            \If {\Cref{cond} is not False}
                \State $w = w + 1$
            \EndIf
        \EndWhile
	\end{algorithmic} 
\end{algorithm}

\section{Experiments}
\label{Experiments}

\paragraph{Datasets.} 
With 10 years of regional half-hourly electricity consumption data and the current consumption levels for the regions, we can aim to predict the short-term electricity needs of France over the next three hours. From a set of time series with similar characteristics, the goal is to predict a hold-out time series from GPs trained on the rest of the set.
As a proof of concept, in the present paper we experiment on artificially generated periodic data. The synthetic data is generated with trend, periodic and noise components,
using the \texttt{mockseries} Python package \url{https://mockseries.catheu.tech/} which allows us to create time series indexed to a chosen time frame, and with constraints.

\paragraph{Evaluation.} 
We evaluate using the Median Absolute Error (MAE) as it is robust to outliers whilst giving an error in the same unit as the output. Given $y_i$ the $i^{th}$ sample and $\hat{y}_i$ its predicted value, the MAE is calculated by: $ \mathrm{MAE}(y, \hat{y}) = \mathrm{median} \bigl({|y_i - \hat{y}_i|}_{i=1}^I\bigr) $.
We use MAGMA to train GPs on the data. We evaluate the model, as well as the \textsc{Domino} trained on the model's samples. The test data is evaluated following the protocol in \Cref{appendix:evaluation}. 

\paragraph{Ablation studies.} We study the impact of hyperparameter adjustment in \textsc{Domino}. We experiment with time series length to find the optimal number of points per time series for MAGMA and \textsc{Domino}, and we establish the relative performance of both algorithms. Our code and data are available online at \url{https://anonymous.4open.science/r/domino-effect-155D/}.


\paragraph{Results.} 
We gather in \Cref{tab:results_length} the results of performance when varying the time series length for MAGMA and \textsc{Domino}; the cross-validation results are in \Cref{tab:results_cv}. \textsc{Domino} consistently outperforms MAGMA by a significant margin.

\begin{wraptable}{l}{0.6\textwidth}
   \centering
   \caption{Average (std) MAGMA and \textsc{Domino} MAE on 10 runs.}
    \begin{tabular}{>{\centering\arraybackslash}m{2cm} >{\centering\arraybackslash}m{2cm} >{\centering\arraybackslash}m{2cm}}
       \toprule
       \textbf{Length $N$} & \textbf{MAGMA} & \textbf{\textsc{Domino}} \\
       \midrule
       50 & 8.089 & 4.405 \\
          & (0.015) & (1.006) \\ \hline
       100  & 6.059 & 4.526 \\
        & (0.085) & (0.932) \\ \hline
       150 & 33.454 & 3.618 \\
          & (0.108) & (0.559) \\ \hline
       200 & 48.108 & \textbf{3.524} \\
          & (0.113) & (0.386) \\ \hline
       250 & 5.991 & 4.511 \\
          & (0.029) & (0.336) \\
       \bottomrule
   \end{tabular}
   \label{tab:results_length}
\end{wraptable}

\paragraph{Discussion and limitations.}
We have used a uniform probability distribution on the sampled time series but can extend the work to a scenario with prior knowledge and therefore a known probability distribution across the samples at initialisation.

\begin{wraptable}{l}{0.6\textwidth}
   \centering
   \caption{Average (std) MAGMA and \textsc{Domino} MAE at cross-validation on 10 runs.}
   \begin{tabular}{>{\centering\arraybackslash}m{2cm} >{\centering\arraybackslash}m{2cm} >{\centering\arraybackslash}m{2cm}}
       \toprule
       \textbf{Length $N$} & \textbf{MAGMA} & \textbf{\textsc{Domino}} \\
       \midrule
       50 & 8.91 & 4.114 \\
          & (0.319) & (0.419) \\ \hline
       100  & 6.624 & 5.058 \\
        & (0.363) & (0.333) \\ \hline
       150 & 33.973 & 4.708 \\
          & (0.466) & (0.334) \\ \hline
       200 & 48.412 & 9.679 \\
          & (0.326) & (0.079) \\ \hline
       250 & 56.215 & 4.608 \\
          & (0.342) & (0.226) \\
       \bottomrule
   \end{tabular}
   \label{tab:results_cv}
\end{wraptable}

\textsc{Domino} dramatically improves on the MAGMA algorithm. A natural next step is to conduct a similar study on MAGMAClust \citep{leroy_cluster-specific_2023}, a generalisation of MAGMA which learns cluster-specific means and infers clusters whilst learning the common means.

MAGMA can handle covariates. This is a natural next step for the \textsc{Domino} algorithm. 
We have worked with inputs on a regular grid, as there is a very regular data stream for electricity consumption and we worked with similar data. This approach will be limited where there is an irregular input and calls for an adaptation of \textsc{Domino}.

\paragraph{Hyperparameters.} \textsc{Domino} is controlled by hyperparameters which determine the optimal maximal number of training epochs (30), the percentage of the minimum-maximum range of the output which is an acceptable KL divergence $\delta$ between the training time series and the random walk learned (5\%), the maximal number of points (all outputs, all time series combined) which can be over the $\delta$ value (3\%) and the $\lambda$ regularisation parameter which smooths the weights when calculating the probability of each time series for the next sample (0.5). The full set of ablation studies are detailed in \Cref{appendix:hps}.

\paragraph{Conclusion.}
In this work, we have used the MAGMA algorithm to predict short-term electricity usage based on GPs with common means. We then performed a random walk on samples of these GPs, which is iterated until there is a low divergence between the sampled time series and the points of the random walk. Our experiments show that this approach, called \textsc{Domino}, yields superior predictive results on a synthetic dataset. This is very promising to tackle similar problems in sparse data settings, with less computational resources, or heavy data settings, paving the way to more frugal probabilistic settings.

\bibliography{bdu_2024}

\appendix

\section{Evaluating the \textsc{Domino} algorithm}
\label{appendix:evaluation}

With the time series on which \textsc{DOMINO} has been trained as well as their weights, the model is queried by inputting a set of $M$ time points such that $M < N$. The time points contain the same information as the training data, that is either $\mathbf{t}$ as an input; and $\mathbf{y}$ is the output. Made up of $M$ time-steps, these have dimension $\mathbf{y}, \mathbf{t} \in \mathbb{R}^M$.

The \textsc{Domino} algorithm is also given a set of input time points over which an output must be returned - these will be made up of $\mathbf{x} \in \mathbb{R}^N$ and will consist of the first $M$ points from the query input, plus $N-M$ extra points which will be used to predict the next points.

The random walk, starting at the $M+1^{th}$ point, uses the training time-series and their probabilities to predict the rest of the time series.

\section{Hyperparameter tuning}
\label{appendix:hps}

The \textsc{Domino} model has multiple hyperparameters, which control the learning of probabilities for underlying individuals. We run ablation studies for each hyperparameter: the maximal number of epochs for the learning (\Cref{tab:hps_max_epochs}), the maximum percentage $\delta$ of the data range which is a possible divergence threshold between the \textsc{Domino} and the underlying training individuals (\Cref{tab:delta}), the maximum percentage of points in all the time series which can be over the $\delta$ hyperparameter (\Cref{tab:max_over_delta}), and the regularisation parameter $\lambda$ (\Cref{tab:lambda}). The best results for the hyperparameter are given in bold.

\begin{table}[ht!]
   \centering
   \begin{minipage}{0.45\textwidth}
        \centering
        \caption{Hyperparameter tuning: average (std) MAE for maximal number of epochs on 10 runs.}
        \begin{tabular}{>{\centering\arraybackslash}m{2cm} >{\centering\arraybackslash}m{2cm} >{\centering\arraybackslash}m{2cm}}
           \toprule
           \textbf{Max epochs} & \textbf{Result} & \textbf{CV} \\
           \midrule
           5 & 5.750 & 5.302 \\
           & (0.753) & (0.313) \\ \hline 
           10 & 5.648 & 5.244 \\
           & (0.792) & (0.225) \\ \hline 
           15 & 5.082 & 5.314 \\
            & (0.792) & (0.308) \\ \hline 
           20 & 5.460 & 5.174 \\
            & (0.668) & (0.331) \\ \hline 
           25 & 5.206 & 4.964 \\
            & (0.544) & (0.215) \\ \hline 
           30 & \textbf{5.049} & 5.054 \\
            & (0.409) & (0.251) \\
           \bottomrule
       \end{tabular}
   \label{tab:hps_max_epochs}
   \end{minipage} \hfill
   \begin{minipage}{0.45\textwidth}
        \centering
        \caption{Hyperparameter tuning: average (std) MAE for $\delta$ threshold for divergence on 10 runs.}
        \begin{tabular}{>{\centering\arraybackslash}m{2cm} >{\centering\arraybackslash}m{2cm} >{\centering\arraybackslash}m{2cm}}
           \toprule
           \textbf{$\delta$} & \textbf{Result} & \textbf{CV} \\
           \midrule
           1\% & 5.413 & 5.247 \\
           & (0.597) & (0.341) \\ \hline 
           2\% & 5.218 & 5.034 \\
           & (0.502) & (0.218) \\ \hline 
           3\% & 5.187 & 5.211 \\
            & (0.537) & (0.307) \\ \hline 
           5\% & \textbf{5.185} & 5.096 \\
            & (0.583) & (0.283) \\ \hline 
           10\% & 5.640 & 5.331 \\
            & (0.594) & (0.453) \\
           \bottomrule
       \end{tabular}
   \label{tab:delta}
   \end{minipage}
\end{table}

\begin{table}[ht!]
   \centering
   \begin{minipage}{0.45\textwidth}
        \centering
        \caption{Hyperparameter tuning: average (std) MAE for maximal number of values over $\delta$ on 10 runs.}
        \begin{tabular}{>{\centering\arraybackslash}m{2cm} >{\centering\arraybackslash}m{2cm} >{\centering\arraybackslash}m{2cm}}
           \toprule
           \textbf{Maximum percentage of values over $\delta$} & \textbf{Result} & \textbf{CV} \\
           \midrule
           1\% & 5.130 & 4.67 \\
           & (0.732) & (0.360) \\ \hline 
           2\% & 5.249 & 5.102 \\
           & (0.434) & (0.297) \\ \hline 
           3\% & \textbf{5.130} & 5.156 \\
            & (0.637) & (0.241) \\ \hline 
           5\% & 5.344 & 5.110 \\
            & (0.521) & (0.375) \\ \hline 
           10\% & 5.275 & 5.054 \\
            & (0.499) & (0.190) \\
           \bottomrule
       \end{tabular}
   \label{tab:max_over_delta}
   \end{minipage} \hfill
   \begin{minipage}{0.45\textwidth}
        \centering
        \caption{Hyperparameter tuning: average (std) MAE for $\lambda$ the regularisation parameter on 10 runs.}
        \begin{tabular}{>{\centering\arraybackslash}m{2cm} >{\centering\arraybackslash}m{2cm} >{\centering\arraybackslash}m{2cm}}
           \toprule
           \textbf{$\lambda$} & \textbf{Result} & \textbf{CV} \\
           \midrule
           0.5 & \textbf{4.698} & 5.450 \\
           & (0.473) & (0.407) \\ \hline 
           1 & 5.308 & 5.231 \\
           & (0.506) & (0.566) \\ \hline 
           1.5 & 5.315 & 5.129 \\
            & (0.360) & (0.540) \\
           \bottomrule
       \end{tabular}
   \label{tab:lambda}
   \end{minipage}
\end{table}

\end{document}